\documentclass[conference]{IEEEtran}

\usepackage{cite}

\ifCLASSINFOpdf
 \usepackage[pdftex]{graphicx}
 \graphicspath{figures}
 \DeclareGraphicsExtensions{.pdf,.jpeg,.png}
\else
\fi

\usepackage{tikz}
\usepackage{amsmath,amsfonts,amssymb}
\usepackage{colortbl}
\usepackage{comment}
\usepackage{multirow}
\usepackage{makecell}
\usepackage{url}
\usepackage{algorithm}
\usepackage{algorithmicx}
\usepackage{textcomp}
\usepackage{hyperref}
\usepackage[caption=false,font=footnotesize]{subfig}
\usepackage{booktabs}
\usepackage{tabularx}
\usepackage{stfloats}
\begin{document}

\title{What's Cracking? A Review and Analysis of Deep Learning Methods for Structural Crack Segmentation, Detection and Quantification
}

\author{Jacob K\"onig,
        Mark David Jenkins,
        Mike Mannion,
        Peter Barrie,
        and Gordon Morison
\thanks{The authors are with the School of Computing, Engineering and Built Environment, Glasgow Caledonian University, G4 0BA Glasgow, Scotland (Email: jacob.koenig@gcu.ac.uk).}}

\markboth{Journal of \LaTeX\ Class Files,~Vol.~14, No.~8, August~2015}%
{Shell \MakeLowercase{\textit{et al.}}: Bare Demo of IEEEtran.cls for IEEE Journals}

\maketitle
\thispagestyle{plain}
\pagestyle{plain}

\begin{abstract}
Surface cracks are a very common indicator of potential structural faults. Their early detection and monitoring is an important factor in structural health monitoring. Left untreated, they can grow in size over time and require expensive repairs or maintenance.
With recent advances in computer vision and deep learning algorithms, the automatic detection and segmentation of cracks for this monitoring process have become a major topic of interest. 
This review aims to give researchers an overview of the published work within the field of crack analysis algorithms that make use of deep learning. It outlines the various tasks that are solved through applying computer vision algorithms to surface cracks in a structural health monitoring setting and also provides in-depth reviews of recent fully, semi and unsupervised approaches that perform crack classification, detection, segmentation and quantification. Additionally, this review also highlights popular datasets used for cracks and the metrics that are used to evaluate the performance of those algorithms. Finally, potential research gaps are outlined and further research directions are provided.
\end{abstract}

\section{Introduction}
The public infrastructure of today's society is largely made up of structures built with concrete, such as bridges and tunnels or pavements that can consist of a variety of materials such asphalt, concrete or different kinds of stones.
With increased use and aging structures, the need for maintenance increases, which when not addressed properly, can lead to poor conditions or structural deficiencies. This is a common problem. For example, in the US a recent report on the state of its infrastructure from the American Society of Civil engineers states that on average every fifth mile of highway is of poor condition as well as over 2000 dams and over 46000 bridges being structurally deficient \cite{2020COVID19Impacts}. In the UK, a report of the BBC stated that a total of about 10\% of road networks under maintenance of the local authorities are of improper condition \cite{homer2018ThousandsMiles}.
The process of structural health monitoring (SHM) and assessment is usually carried out in conjunction with those maintenance measures. This is an important step, as here the general state of structures is taken into account and eventual anomalies are cataloged and prioritized. 
However, this process is labor-intensive, as it is often carried out at the target location, at times is dangerous, due to structures being not easily accessible and can lead to downtimes in the use of a structure as bridges, tunnels, or roads that need to be closed off to public access. 
Additionally, as the process is often carried out manually, human factors come into play which may negatively affect the outcome of such structural assessment. Namely, the inspectors may be fatigued, have inconsistent assessments, need specific training, and may provide subjective results \cite{oliveira2009AutomaticRoad}. 
Due to recent developments in regards to unmanned capturing systems with robots or UAVs, some of the data capturing issues can be addressed \cite{pan2018DetectionAsphalta,sony2019LiteratureReview, yuan2021NovelIntelligent, jang2021AutomatedCrack}. Nonetheless, even after data has been captured it still needs to be analyzed and evaluated. 

\begin{figure}
    \centering
    \includegraphics[width=\linewidth]{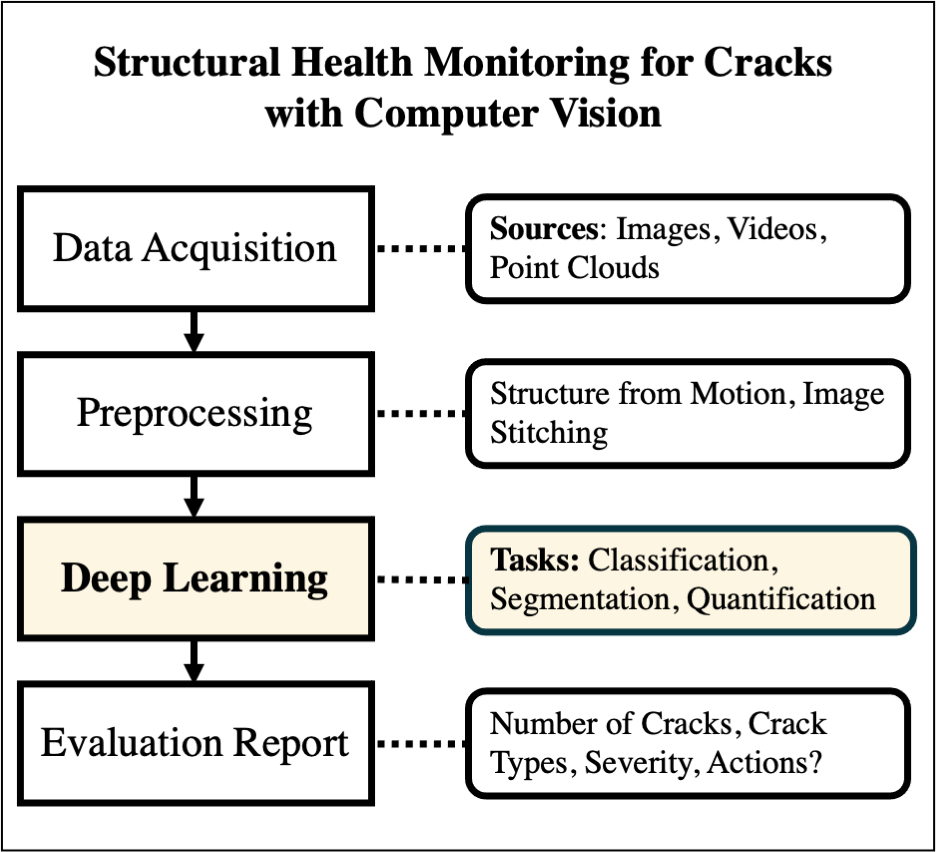}
    \caption{Deep learning based computer vision algorithms for cracks in the context of the structural health monitoring}
    \label{fig:context}
\end{figure}

During the SHM process, the goal is to find indicators for current or future damage. Some common damage types that appear on structures are surface cracks on pavements, concrete, bricks, stone and asphalt, spalling of concrete and cracked or corroded steel \cite{spencerjr2019AdvancesComputer}. 
As one of the most common defects, cracks can reduce the structural integrity of surfaces as well as present early indicators of further structural fatigue. Therefore, early detection of them is a vital step to keep infrastructure and pavements safe and operational as long as possible. 
Most of the techniques that are concerned with the finding of cracks include non-destructive techniques, meaning the surfaces are not damaged during their evaluation. Those techniques may include simple visual examinations or image/video capturing \cite{sony2019LiteratureReview}, creating 3D representations using lidar scanners \cite{riveiro2016AutomatedProcessing, song2018ApplicationFinite} as well as examinations using ultrasonic waves\cite{mutlib2016UltrasonicHealth, erdogmus2020NovelStructural}. It is clear, that after the data has been captured using those techniques, it can then be revisited at a later stage and anomaly detection algorithms can be run. This has inspired a large body of research in the last decades which has studied the automatic detection of cracks as surface faults, ranging over a variety of data types. 
When using 2-dimensional image or video data or 3-dimensional data in the form of point-clouds, the detection of cracks can present to be quite a difficult task; Cracks may often not appear in uniformly, they may be of different shapes and sizes as well as obstructed through leaves, moss or otherwise be partially hidden by occlusions. Due to some cracks also being extremely small, it can also be hard to distinguish between them and their background. Whilst these conditions may be difficult for humans to work with, a possible solution can be found in the form of machine learning (ML). 
ML algorithms in the computer vision domain have seen a great increase in their performance over the last decades, partly enabled by the influx of much more powerful hardware and software. These advances in the field of ML, combined with computer vision applications, have driven a large number of technologies and applications in recent years. This ranges from enabling of automated driving \cite{milz2018VisualSLAM, rashed2021GeneralizedObject}, to surpassing human performance on image classification \cite{he2015DelvingDeep} or aiding specialists in the medical field with the diagnosis of different scans such as CT \cite{zhou2020RapidAccurate} or X-Rays \cite{baltruschat2019ComparisonDeep}. A majority of state-of-the-art methods in those tasks are driven by deep neural networks, which belong to the field of deep learning (DL) a subset of ML. 
This has also enabled a surge in research which is concerned with the automation of parts of the surface health inspection process by automatically analyzing and quantifying cracks as surface faults \cite{riid2019PavementDistress, louk2020PavementDefect, li2021AutomaticDefect, jang2021AutomatedCrack}. The general process for the integration of DL into the structural health monitoring context for cracks is shown in \autoref{fig:context}.

Recently, there has been an influx in publications of reviews and surveys that are concerned with the automation of the SHM process and the integration of ML. 
General applications of DL in construction are outlined in \cite{khallaf2021classification} and recent advances in computer vision methods including ML applied to the SHM process in \cite{spencerjr2019AdvancesComputer, dong2020ReviewComputer}. In \cite{gopalakrishnan2018DeepLearninga, cao2020ReviewPavementa, lu2020advances} image processing methods exclusively for pavements including ML and DL are reviewed. Thorough reviews are provided for a wide range of image processing and ML techniques for surface cracks \cite{zakeri2017ImageBased, mohan2018CrackDetectiona}. However, they also include a majority of the non-DL based ML methods. Whilst some other works like in \cite{spencerjr2019AdvancesComputer, munawar2021image, du2021ApplicationImage, cao2020ReviewPavementa, khallaf2021classification, dong2020ReviewComputer} include sections about DL methods for surface cracks, it is not a major focus, therefore, leaving out specific details about the different DL based approaches. Some works focus exclusively on surface cracks and DL \cite{lu2020advances, gopalakrishnan2018DeepLearninga, hsieh2020MachineLearning}. Specifically, \cite{gopalakrishnan2018DeepLearninga} investigates DL for pavement distress images, however since its initial publication several other different methods have been released. \cite{lu2020advances} goes into detail about various datasets and some example DL models, but missing out on methods that perform crack quantification and use other learning types than supervised learning. Most closely related to our work is \cite{hsieh2020MachineLearning} outlining some tasks, metrics and models for ML in surface cracks. However this work does not focus on the different kinds of learning types, rather implements fully supervised learning methods and outlines their limitations. 
Therefore, to address some of the shortcomings of those previous surveys we exclusively focus on DL and review a significant number of recent works which are concerned with DL and their target tasks of cracks classification, detection, segmentation quantification. We aim to focus on techniques that make use of 2-D data and split them by their approach of ML, which is based on the detail for the labels of the data. Previous works did not have this as the main focus and have also lacked a focus on quantification. Additionally, we then outline the approaches, categorized by their tasks performed and methods of applications on cracks. Furthermore, we also include an overview of common metrics used to assess performance for algorithms in this field, as well as outline some of the most popular benchmark datasets that are used. Finally, we examine potential research gaps, provide an overview of the issues with the research in this domain, and include an outlook into further potential future work. Therefore the main contributions and differences to other reviews of this paper can be summarized as follows:
\begin{enumerate}
    \item An in-depth review only focusing on DL based relevant models and applications, specifically for structural surface cracks using computer vision, is provided. This is split into the main tasks that are applied to cracks of classification, detection, segmentation and quantification, as well as the type of learning which dictates the data labels that are required. 
    \item A thorough overview of common structural crack datasets used for training and testing of models as well as the evaluation metrics that are used to compare the performance of DL based models is provided. Additionally, the shortcomings of current datasets and metrics are outlined as well. 
    \item An outlook into future research and challenges that are faced by approaches implementing DL for structural cracks
\end{enumerate}

This work specifically chooses to omit a detailed comparison of the performance of the methods, as the approaches differ too much in the choice of datasets (and their respective train/test splits) as well as metrics used.

\section{Tasks}

\begin{figure}[!t]
    \centering
    \setlength{\tabcolsep}{1pt}
    \renewcommand{\arraystretch}{0.8}
    \newcolumntype{T}{>{\raggedright\arraybackslash} m{0.04\linewidth}}
    \newcolumntype{Y}{>{\raggedright\arraybackslash} m{0.05\linewidth}}
    \newcolumntype{B}{>{\centering\arraybackslash} m{0.4\linewidth}}
    \begin{tabular}{TYBB}
            & & Input & Output \\ 
            & \rotatebox[origin=c]{90}{(a) Classification}
            & \includegraphics[width=\linewidth]{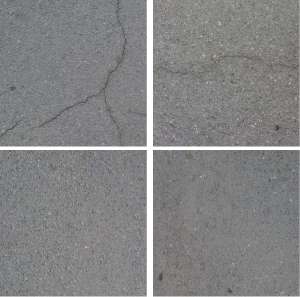}
            & \includegraphics[width=\linewidth]{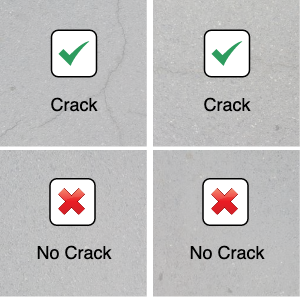} \\
            & \rotatebox[origin=c]{90}{(b) Detection}
            & \includegraphics[width=\linewidth]{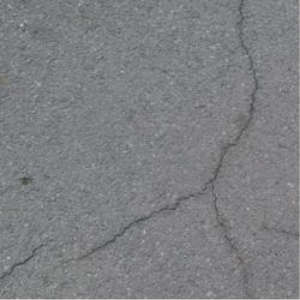}
            & \includegraphics[width=\linewidth]{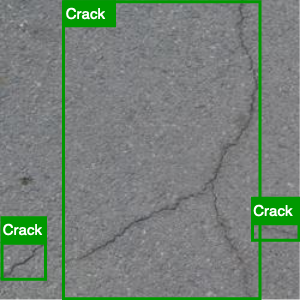} \\            
             \rotatebox[origin=c]{90}{(c) Semantic}
            &\rotatebox[origin=c]{90}{Segmentation}
            & \includegraphics[width=\linewidth]{figures/tasks/input.png}
            & \includegraphics[width=\linewidth]{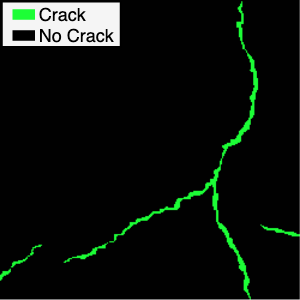} \\   
            \rotatebox[origin=c]{90}{(d) Instance}
            &\rotatebox[origin=c]{90}{Segmentation}
            & \includegraphics[width=\linewidth]{figures/tasks/input.png}
            & \includegraphics[width=\linewidth]{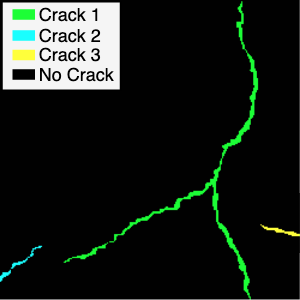} \\   
    \end{tabular}

\centering
\caption{Simplified overview over common computer vision tasks when performing SHM for cracks using ML techniques on images. (a) Single class image classification, determining the presence of a crack. (b) Detection encapsulates crack regions in bounding boxes. (c) Semantic segmentation, where all pixels belonging to the crack class are labeled, and (d) Instance segmentation, which also differentiates between instances of cracks on a pixel level.}
\label{fig:tasks_general}
\end{figure}

The process of extracting information about surface cracks using computer vision techniques can be split into different tasks. These tasks align with the classical computer vision tasks of classification, detection and segmentation and their choice is dependant on the amount of detail that is required in the SHM process. Example tasks are shown in \autoref{fig:tasks_general}. The following section outlines the main tasks in a SHM setting that are applied to surface cracks using computer vision:

\subsection{Crack Classification} 

\begin{figure}[!t]
    \centering
\captionsetup[subfigure]{labelformat=empty}
\subfloat[Longitudal Crack]{\includegraphics[width=\linewidth]{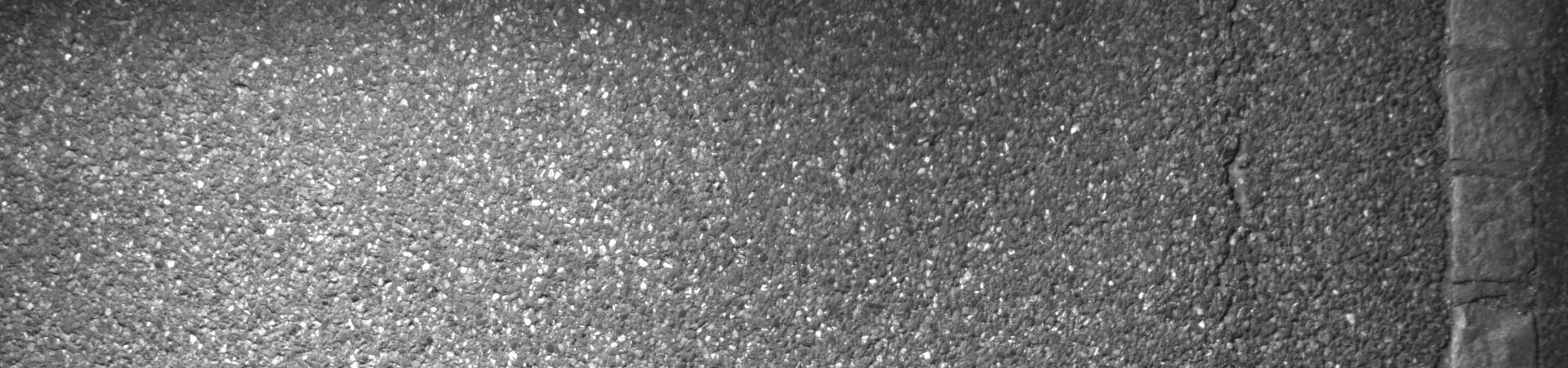}} \\ \vspace{-0.03\linewidth}
\subfloat[Transverse Crack]{\includegraphics[width=\linewidth]{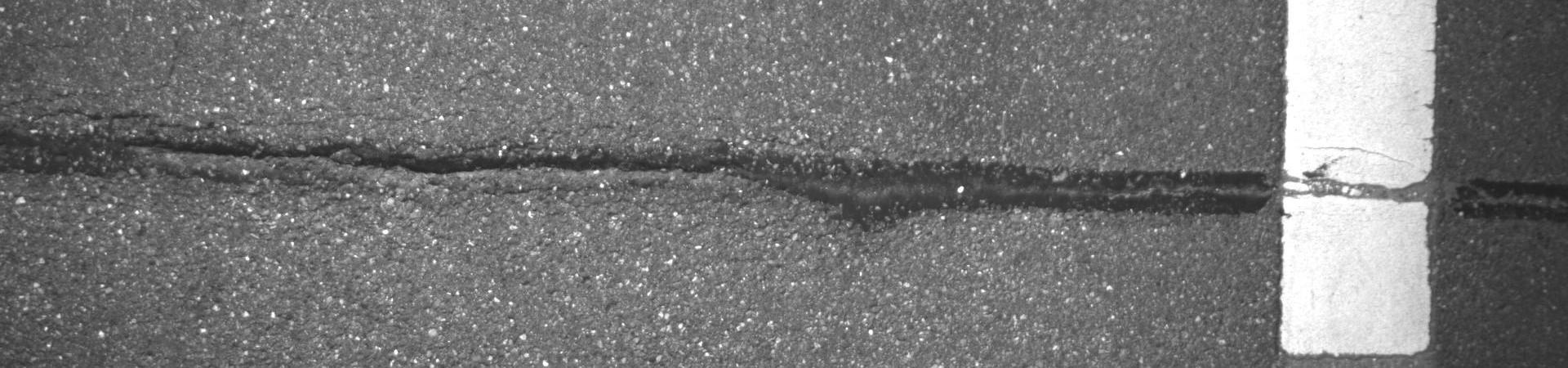}} \\ \vspace{-0.03\linewidth}
\subfloat[Alligator Crack]{\includegraphics[width=\linewidth]{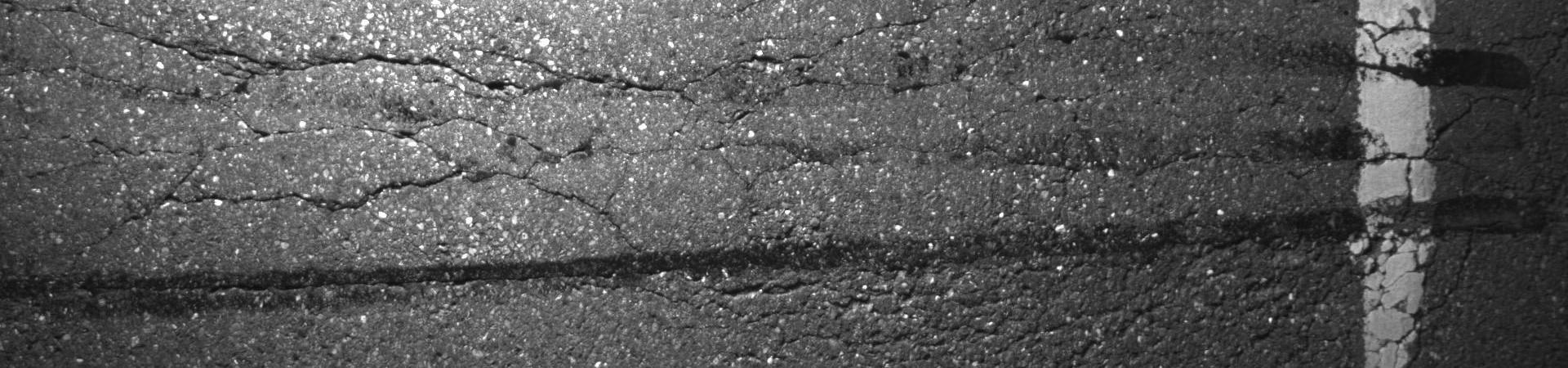}} \\ 
    \caption{Examples of different crack-types on road surfaces \cite{eisenbach2017HowGet}.}
    \label{fig:crack_types}
\end{figure}

\begin{figure}
\centering
\captionsetup[subfigure]{labelformat=empty}
\subfloat[Input]{\includegraphics[width=0.32\linewidth, ]{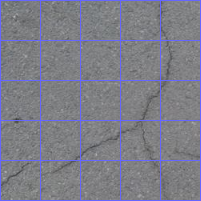}}\vspace{-0.02cm}
\subfloat[Patch Classification]{\includegraphics[width=0.32\linewidth]{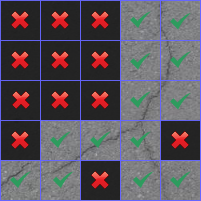}}\vspace{-0.02cm}
\subfloat[Coarse Detection]{\includegraphics[width=0.32\linewidth]{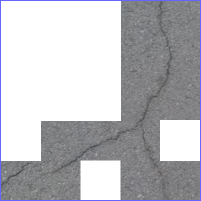}}\vspace{-0.02cm} \\
    \caption{Localisation of crack regions in an image by classifying whether individual patches contain a crack. This yields a coarse detection map of crack-regions.}
    \label{fig:splitcrack}
\end{figure}

In the classical context for computer vision, the goal of the image classification task is to assign the correct class label onto an image. 
For classification within the space of cracks in SHM, the simplest task is to determine whether an image contains a 
crack or if it is an image without one. However, even though this is only a binary classification task, it may be challenging as there may be other surface defects that appear similar to cracks. Expanding from this two-label classification task, the crack classification task can also be extended to not only determine if there is a crack present but also deducing the crack type. Some examples of different crack types on road surfaces are shown in \autoref{fig:crack_types}, with their type being determined by their direction and pattern.
In the context of a general classification method for SHM, cracks may not be the only surface anomaly that has been labeled. For example, road surfaces have other anomalies such as potholes and road patches \cite{eisenbach2017HowGet}.
In terms of obtaining the classification label, in comparison with the other tasks, the amount of work required to label the data is minimal. 
This task can generally be categorized as the simplest task, as further tasks not only highlight the presence of a crack but also provide some additional information such as the location of the cracks. However, some methods exploit algorithms that have been trained only for crack classification to provide some form of localization. This is achieved by splitting the data into sub-regions such as image patches and performing classification on each patch, followed by aggregating the patches. This can lead to more localized results, such that the results may be that of crack detection. An example of this is shown in \autoref{fig:splitcrack}.

\subsection{Crack Detection}
Crack detection expands on classification and does not only determine the existence of cracks but aims to provide further information by highlighting the location of the crack. This can be beneficial as basic classification tasks merely emphasize the existence of cracks but still leave it up to the observer to find the actual crack, whereas the detection task aims to provide some kind of localization of cracks. When classifying sub-regions of an image and stitching them back together to a larger piece, one can approximate the location of a crack. Another type of detection is the creation of bounding boxes. This is a commonly researched task in other large datasets such as COCO \cite{lin2014MicrosoftCOCO} and has gathered quite a large amount of recent work. However, for cracks, this may not always be optimal. As some cracks appear as a single linear structure, in the worst case the bounding box marks only the outermost points of a crack with a large amount of the region inside this box not containing any cracks. 

\subsection{Crack Segmentation}
Crack Segmentation solves the problem of not having sufficient localization of cracks which is detrimental in detection and classification approaches.  The goal of this task is to assign pixels in an image, or voxels in 3-D-volumes a specific class label. The segmentation task can further be broken down into semantic and instance segmentation. The former assigns all pixels or voxels in a piece of input data a class but does not separate between instances. In other words, in an image, multiple cracks would all be labeled as belonging to the class crack. Instance segmentation on the other hand does differentiate between each instance. Whilst different cracks would be assigned the class label - crack, their instance label would differ. An example of this is shown in \autoref{fig:tasks_general}. Additionally, we can also differentiate between binary- and multi-class segmentation. Many works in crack segmentation only aim to provide a binary segmentation of data, as in \cite{zou2018DeepCrackLearning,liu2019DeepCrackDeepa, yang2019FeaturePyramid, konig2021OptimizedDeep, guo2021BARNetBoundary} where there are only two classes, crack and no-crack. In a multi-class segmentation context, cracks and background are only two pixel class-labels next to others, such as in \cite{hoskere2020MaDnetMultiTask} where other structural anomalies are segmented as well such as spalling, corrosion, or exposed rebar.

\subsection{Other Tasks}
Other tasks with which the state of structural cracks can be assessed may include quantification analysis such as determining the \textit{rotation} or providing \textit{measurements} of cracks. Rotation of cracks may be important because maintenance measures may only need to be applied if the crack has a certain orientation \cite{inoue2019DeploymentConsciousa}. However, only a few DL based works cover the orientation prediction of road cracks \cite{inoue2019DeploymentConsciousa, tran2020OneStage}. 

Calculating the measurements of cracks gives information on their size and can also be important for SHM. Maintenance or repair processes may require a crack to exceed a certain size before they are classed as significant and require attention. Additionally, crack measurements can also give some indicators over their severity, with thin cracks not being severe and thicker cracks being the most severe. However, even without performing measurements, crack quantification can also be performed by extending the classification task and not only labeling the presence of cracks but categorizing them into their severity using classes such as [low, medium, high].

Further tasks that can make use of the results of segmentation and detection are \textit{change detection} and \textit{forecasting}. In the former, cracks are analyzed at multiple time steps to show changes in their size and severity. This task however requires data from the same location at different time points which can be difficult to acquire. Using data from different time steps, another task is to forecast the development of surface cracks. Unfortunately, little has been reported on this particular task within the domain of surface cracks. A reason for this may be due to the lack of available open data of cracks over a longer period.

\section{Machine and Deep Learning for Structural Cracks}
The automated detection and segmentation of surface cracks have been a topic of interest for a large amount of time now \cite{luxmoore1973HolographicDetection, mendelsohn1987AutomatedPavement, tanaka1998CrackDetection}.
Early research within this field made use of methods that do not require learning from data but rather employ hand-crafted features. Those included works using mathematical morphology \cite{tanaka1998CrackDetection}, thresholding \cite{oliveira2009AutomaticRoad, peng2015ResearchCrack}, applying Gabor filters \cite{salman2013PavementCrack} and wavelets \cite{wang2007WaveletBasedPavement} or using edge detection algorithms such as Canny \cite{zhao2010ImprovementCanny}. Additional work then also started to focus on common ML algorithms such as random forests \cite{shi2016AutomaticRoad} or support vector machines \cite{fernandes2014PavementPathologies} which also showed promising results and proved to have better adaptability to more diverse data in comparison with handcrafted features.

In 2012, a convolutional neural network was the first DL based algorithm that won the large-scale ImageNet image classification challenge \cite{krizhevsky2012ImageNetClassification}. This breakthrough lead to all successive winners of this challenge also being DL based. Although CNN's and neural networks were used previously, this seemed to inspire many researchers that were starting to use DL for their applications and it spawned many new algorithms. Those CNN based methods showed a much superior performance than the aforementioned ML approaches and quickly became a standard for the tasks within this domain and in general computer vision-based SHM \cite{spencerjr2019AdvancesComputer}. CNN based methods have shown much better capabilities in this task due to their ability to automatically learn high- and low-level features alike. Due to this, a very large majority of the works recently being published in this field make use of DL with other ML methods only contributing to a minority of the work published.

\section{Review of Deep Learning based Methods}

 \begin{table*}[ht]
\renewcommand{\arraystretch}{1.2}
\centering
\caption{Overview and shared properties of the commonly used approaches for surface crack classification (C), detection (D), segmentation (S) and quantification (Q)}
\label{tab:methods}
\begin{minipage}{\linewidth}
\begin{tabularx}{\linewidth}{llXl}
\toprule
\multicolumn{1}{c}{\textbf{Learning}} &
\multicolumn{1}{c}{\textbf{Task}} &
\multicolumn{1}{c}{\textbf{Description}} &
\multicolumn{1}{c}{\textbf{Examples}} \\ \midrule

\multirow[t]{13}{*}{Supervised} 
  & C &  Classification CNN to classify whether images/patches contain a cracks or other defect types. &
  \cite{zhang2016RoadCracka, pauly2017DeeperNetworksa, eisenbach2017HowGet, stricker2019ImprovingVisual} \\ \cline{2-4}
  & D & Classification of patches using a CNN followed by merging to obtain a coarse detection map. & \cite{wang2017GridBasedPavementa, konig2021WeaklySupervisedSurface} \\ 
  \cline{2-4}
  & S  & CNN that classifies the presence of a crack within single pixel. Using a sliding window approach, every pixel within an image is classified.  & \cite{pauly2017DeeperNetworksa, fan2018AutomaticPavementa, inoue2019DeploymentConsciousa} \\
  & S  & Approaches using a standard or slightly modified U-Net  & \cite{cheng2018PixelLevelCrack, jenkins2018DeepConvolutionala, konig2019ConvolutionalNeurala, konig2019SegmentationSurfacea, wu2020CrackDetecting, augustauskas2020AggregationPixelWise} \\
   & S  & Approaches using a U-Net shape with popular image classification architectures as the encoder and a custom decoder   & \cite{lau2020AutomatedPavement, wang2020ConvolutionalNeural, konig2021OptimizedDeep, qu2021CrackDetection} \\
  & S  & Specialised architectures for cracks that are only loosely based on other encoder-decoder approaches  & \cite{zou2018DeepCrackLearning, liu2019DeepCrackDeepa, yang2019FeaturePyramid ,guo2021BARNetBoundary, chen2021EffectiveHybrid, han2021CrackWNetNovel} \\
  & S  & Use of specialised loss functions  designed for cracks or similarly shaped structures & \cite{kobayashi2018SpiralNetF1Based, mosinska2018PixelWiseLoss, li2021FastAccurate} \\
  & S  & Use of custom CNN pooling layers designed for cracks & \cite{konig2019SegmentationSurfacea, zhou2021MixedPooling, xu2021PushingEnvelope, han2021CrackWNetNovel} \\
   & S  & Approaches that use an ensemble of CNNs  & \cite{fan2020EnsembleDeep, wang2020NeuralNetwork} \\
  & S  & Approaches that make use of generative models to either generate extra data for algorithm training or use them for segmentation  & \cite{gao2019GenerativeAdversarial, zhang2020CrackGANPavement} \\
  \cline{2-4}
  & Q  & Approaches that use a CNN to predict the size of a crack & \cite{tong2017RecognitionAsphalt, tran2020OneStage} \\
  & Q  & Two-step approaches that first segment a crack followed by calculating their dimensions & \cite{tong2017RecognitionAsphalt, yang2018AutomaticPixelLevel, kim2019ImageBasedConcrete, park2020ConcreteCrack} \\
\midrule
\parbox[t]{2cm}{Semi and Weakly \\Supervised} & S  & Two stage approaches in which a CNN classifier is trained on image labels and its class activation maps are used to create pseudo segmentation labels of the training images. Those pseudo labels are then used to train a segmentation algorithm.   & \cite{dong2020PatchBasedWeakly} \\
& S  & Approaches that train on classification labels and then use thresholding to create a segmentation map & \cite{fan2019RoadCrack, konig2021WeaklySupervisedSurface} \\
& S & Training on coarse segmentation labels   & \cite{inoue2020CrackDetection} \\
\cline{2-4}
& Q & Rotation analysis of cracks by rotating the input   & \cite{inoue2019DeploymentConsciousa} \\
\midrule
\multirow[t]{1}{*}{Unsupervised} 
& S & Transformation of an input image into latent space or frequency domain before reversing transforming it back into an image. The differences between the input and output then segment areas belonging to cracks or other anomalies. & \cite{chow2020AnomalyDetection, yu2020UnsupervisedPixelLevel} \\

\bottomrule
\end{tabularx}
\end{minipage}
\end{table*}
 
For this review, the different methods to tackle crack analysis using DL are split into the different approaches of ML and how they use the available data: supervised, semi as well as weakly supervised, and unsupervised learning.

The most common approach in this domain is to use supervised learning. The goal for this task is to find a mapping function that aims to map an input to output, after having been trained on training input and output data 
The second group of approaches falls into the category of semi and weakly supervised learning. For semi-supervised learning, not all of the data that is being used for training is labeled. Examples for this can include learning to perform semantic segmentation of crack images using an image database where only a subset of data is labeled with segmentation labels and then the rest consisting of basic classification labels. Weakly supervised learning usually consists of learning with labels that are not entirely accurate; In segmentation, some crack regions may be wrongly or not annotated at all. The last group of approaches makes use of unsupervised learning. Here the goal of the algorithm is to try to find patterns but without having labeled data available. 
\autoref{tab:methods} highlights the main approaches and gives an overview of general traits of DL based approaches for the various crack tasks.
The following subsections contain a more detailed breakdown of DL based works for each of those learning approaches.

\subsection{Supervised Learning}
A majority of the works published within in the various tasks for surface cracks make use of supervised learning. For supervised learning, each training input will have a corresponding labeled output. After passing through an input, a prediction is generated by the algorithm which is then compared with the actual output label.
The acquisition of data for this approach can pose a problem for large-scale applications. Depending on the level of detail required in the labels, it can take up to several minutes to label a single image \cite{inoue2020CrackDetection}. This is a domain-wide issue with ML for computer vision often requiring large amounts of labeled data to train effectively. Popular image datasets such as ImageNet and COCO crowdsourced their labeling but resources such as time, money, and lack of expertise might hinder the collation of sufficient labels. Nevertheless, there are several annotated public datasets available and some of the most popular ones are outlined in \autoref{sec:datasets}.

\subsubsection{Supervised Crack Classification}
Common approaches for supervised crack classification train- and test on data that either contain only the classes ``crack" and ``no-crack" or also add additional defect classes that describe other visual anomalies such as potholes or fixed road patches \cite{eisenbach2017HowGet}.
One of the earliest approaches for classification using CNNs is by Zhang \textit{et al.} \cite{zhang2016RoadCracka} where a shallow CNN using 4 convolutional and 2 fully connected layers is trained and tested on small, $99 \times 99$ pixel patches which have been annotated as to whether they contain a crack or not. This was then improved upon in \cite{pauly2017DeeperNetworksa} where it was found that classification networks for cracks with more layers (e.g. deeper networks) improve the classification performance.

\subsubsection{Supervised Crack Detection}
Early approaches for the detection of cracks leverage the results from the previous section, but instead of classifying full images or random patches, images were split into a grid of patches which were then classified individually to provide some localization of cracks. \autoref{fig:splitcrack} shows this kind of approach. A very early work using this technique was released as early as 1993 \cite{kaseko1993NeuralNetworkBased}, where the image is preprocessed using image equalization, thresholding, and noise reduction before feeding the patches into a simple fully-connected neural network to train. This network was then able to determine if - and what type of crack such as longitudinal, transverse or alligator was existent in the patch. 
Another similar approach is shown in \cite{wang2017GridBasedPavementa} where images are split into square patches with a size of 32 or 64 pixels, followed by classification of each patch, this time using a CNN, as to whether they contain a crack. Upon stitching those patches back to the size of the image, an estimated localization of cracks can be obtained. This approach then uses the crack regions within the image in PCA to calculate the eigenvectors and eigenvalues of the crack-skeleton to classify whether cracks are longitudinal, transverse or alligator cracks. 

\subsubsection{Supervised Crack Segmentation}
Supervised crack segmentation requires data that is labeled on a pixel-based level as to whether a crack is present or not. There are a variety of approaches how this can be performed from using a patch-based approach \cite{fan2018AutomaticPavementa, inoue2019DeploymentConsciousa} to an encoder-decoder approach \cite{cheng2018PixelLevelCrack, zou2018DeepCrackLearning, liu2019DeepCrackDeepa}, or combinations thereof \cite{jenkins2018DeepConvolutionala, konig2019ConvolutionalNeurala}. In terms of available research, this domain seems to be the most active with many new works being released.

\begin{figure}
    \centering
    \includegraphics[width=0.8\linewidth]{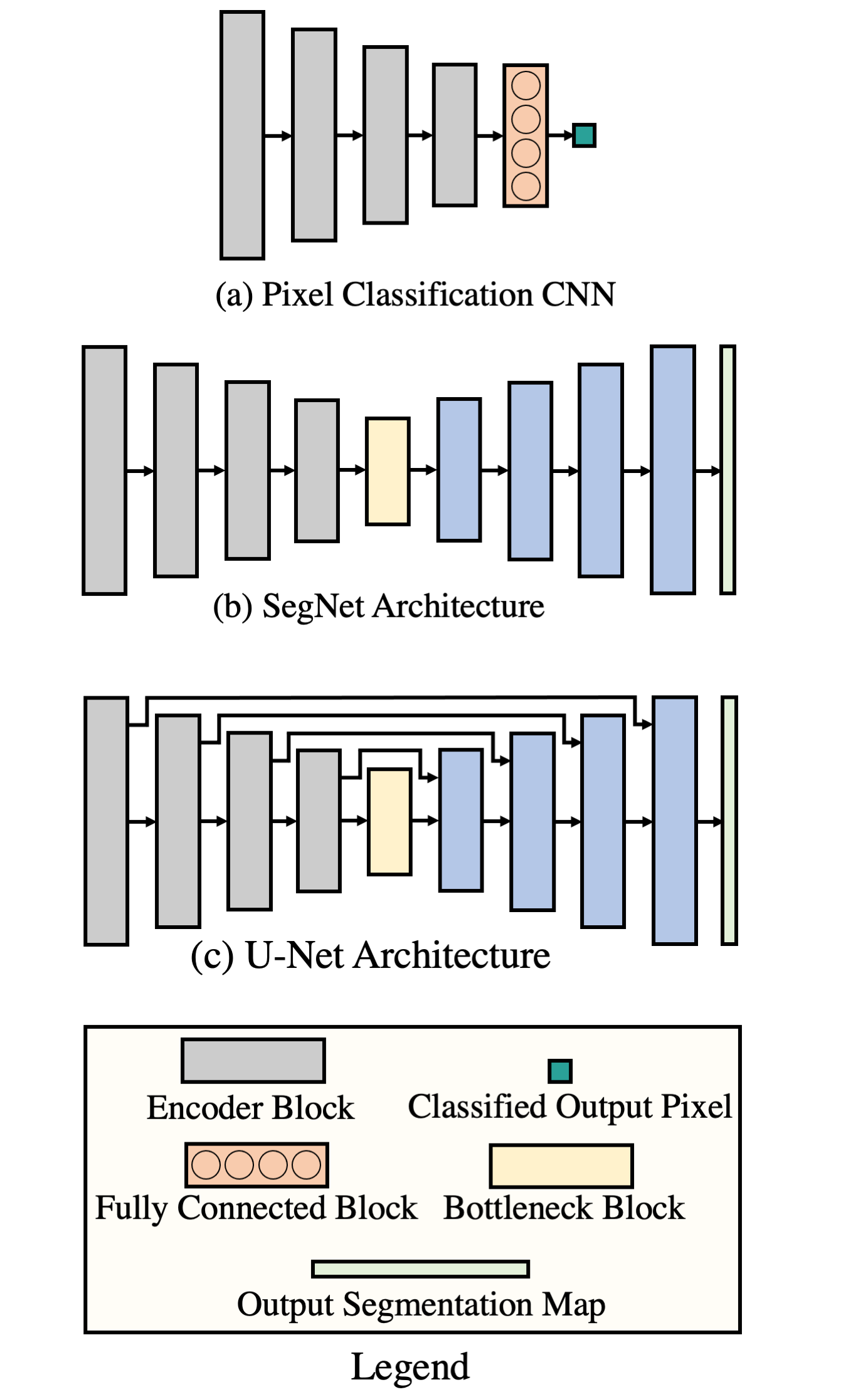}
    \caption{Typical CNN architectures for semantic segmentation often employed for cracks. All blocks contain successions of convolutional, optional batch normalization and activation layers. Within the encoder blocks, the last layer is a pooling layer to downsample the spatial resolution. The first layer in the decoder blocks is an upsampling layer to increase the spatial size. The pixel classification network (a) contains a fully connected element as its last layer, determining whether a \textit{single pixel} belongs to a class. SegNet (b) and U-Net (c) follow an autoencoder-like shape where in U-Net paths feature maps from the encoder are directly passed to the decoder to have better feature retention. The output of (b) and (c) is a segmentation map of the same spatial size as the input. }
    \label{fig:encoder_decoder}
\end{figure}

\textit{Early Approaches:} A very early work \cite{oullette2004GeneticAlgorithm}, even before CNN's for computer vision tasks became immensely popular in this area, used a genetic algorithm approach (instead of standard backpropagation) for selection and evolution of a CNN's weights to segment cracks. The rationale for this was that at the time the genetic algorithm was easier to implement and achieved similar performance to using backpropagation, but it did not appear to gain traction with the rise of modern DL frameworks which are easily available.  
The architectures in \cite{zhang2016RoadCracka, fan2018AutomaticPavementa, inoue2019DeploymentConsciousa} make use of patches to predict a segmentation map and learn using CNN's with fully connected elements at their output. \cite{zhang2016RoadCracka} uses square patches and determines whether the center pixel is within a 5px vicinity of a crack. During testing, a sliding window-based approach is used that considers all possible patch locations and averages their confidence scores to create an output map highlighting crack and non-crack pixels. Fan \textit{et al.} in \cite{fan2018AutomaticPavementa} follow a similar approach but instead of only classifying the center pixel they use a fully connected layer with 25 outputs to generate a segmentation map for a $5 \times 5$ pixel region in the center of a patch. Their testing approach is the same; Using a sliding window approach they feed patches to the trained CNN and outputs are merged to create a segmentation map. A visualization of those  pixel classification CNNs is shown in \autoref{fig:encoder_decoder} (a). 

\textit{Fully Convolutional \& Encoder-Decoder:} Later on, approaches using fully convolutional neural networks gained popularity due to their increased performance. Pioneering approaches here were FCN \cite{long2015FullyConvolutional}, SegNet \cite{badrinarayanan2017SegnetDeep} or U-Net \cite{ronneberger2015UNetConvolutional}. In those approaches, no fully connected layers are used in the last layers to generate the output segmentation map. FCN consists of a succession of convolutions and pooling layers, followed by upsampling of feature maps to generate an output segmentation map of the same spatial size as its input. SegNet follows an autoencoder-like shape - it has an encoder and a decoder part. Within the encoder, features are learned and the spatial size of the feature maps is reduced through pooling whereas in the mirrored decoder the feature maps are resized back to the original size. U-Net follows a similar approach but also adds skip connections between the encoder and the decoder part. This links earlier layers in this architecture with later layers of the same spatial size to reincorporate small features that have been lost due to the down and upsampling process. Whilst those architectures were shown to either perform well on general object segmentation (SegNet and FCN) or medical images (U-Net), following works have also shown their uses within the crack domain. \autoref{fig:encoder_decoder} (b) and (c) show a visualization of SegNet and U-Net respectively. 

Cheng \textit{et al.} \cite{cheng2018PixelLevelCrack} was one of the first approaches to use U-Net on full crack images and showed that it achieved better performance than other ML methods. Later, many more approaches have shown the successful use of U-Net for supervised crack segmentation. In \cite{jenkins2018DeepConvolutionala, konig2019ConvolutionalNeurala, konig2019SegmentationSurfacea} U-Net was trained on small image patches and evaluation was performed using a sliding window approach by averaging the segmentation maps of those patches. Another often used method for segmentation extends U-Net by adding an attention approach to the skip-connections between the encoder and decoder \cite{konig2019ConvolutionalNeurala, wu2020CrackDetecting, augustauskas2020ImprovedPixelLevel}. The attention component creates a spatial attention map, which is combined with the features and has the goal of only passing on features that are relevant for the architecture. Attention components can not only be integrated into the skip connections, they are also used within the blocks that contain convolutions and activations as shown in \cite{wang2020PavementCrack, xiang2020PavementCrack, lau2020AutomatedPavement}. Here, a common approach is to not only redirect the attention to spatial features but it is also used to down-weight channels which do not contain relevant features. 
Another approach is to use an U-Net style architecture but leverage the feature extraction capabilities of popular image classification architectures such as ResNet \cite{he2016DeepResidual} or EfficientNet \cite{tan2020EfficientdetScalable}. Those architectures are used as a U-Net encoder and a custom decoder is added to create the segmentation output. Sample methods that use this are presented in \cite{lau2020AutomatedPavement, wang2020ConvolutionalNeural, konig2021OptimizedDeep}. The work in \cite{konig2021OptimizedDeep} presents that if image classification architectures are used as an encoder, it is beneficial to utilize weights that have been pretrained on datasets such as ImageNet as it provides a stronger feature extraction baseline and increases performance. 
A further addition that has shown to increase performance in those U-Net style architectures is the inclusion of residual style blocks within the U-net as proposed in \cite{konig2019SegmentationSurfacea, ghosh2021CrackWebmodified, konig2021OptimizedDeep}.

Rather than adapting a basic U-Net style architecture, Zou \textit{et al.} \cite{zou2018DeepCrackLearning} create a novel encoder-decoder method by using a custom SegNet-like structure. They show that fusing the output feature maps from multiple scales on both the encoder and decoder achieved superior performance for this task over a basic U-Net or SegNet. 
Another popular method not following the encoder-decoder structure but rather encoder-upscale was introduced by Liu \textit{et al.} \cite{liu2019DeepCrackDeepa}. They used a VGG \cite{simonyan2015VeryDeep} like architecture and, similar to the basic FCN \cite{long2015FullyConvolutional}, upscaled and fused the feature maps to generate the output map with success. 
Qu \textit{et al.} \cite{qu2021DeeplySupervised} propose encoder-decoder architecture that makes extensive use of deep supervision. Their approach uses the DeepLab \cite{chen2018DeepLabSemantic} segmentation architecture as a feature extractor and a novel multi-scale feature fusion which poses to be effective as features from deeper layers are effectively incorporated into the final segmentation output.
A similar approach is proposed in, \cite{qu2021CrackDetection} which uses several of the aforementioned components, such as attention and a residual backbone in combination with fusion of features from multiple levels.

A recent architecture by Guo \textit{et al.} \cite{guo2021BARNetBoundary} proposes a sequential-double-U-Net architecture. The first U-net in this sequence has a parallel edge-detection strand in which a Sobel edge-detection filter is followed by spatial and channel attention modules. The output of this first U-Net is then combined with the edge-detection strand output and passed through the second U-net. The authors show that using this kind of architecture, the boundary information of cracks can be captured more accurately. This is somehow related to the method proposed for cracks in \cite{han2021CrackWNetNovel}, however here the authors utilize a custom pooling function and instead of chaining two U-Nets, add an interim upsampling and downsampling component which is implemented at the bottleneck of the U-Net and increases, then decreases the feature sizes.
To show that for fully-convolutional crack segmentation one does not usually need an encoder-decoder structure, the work in \cite{chen2021EffectiveHybrid} applies convolutional layers with different receptive fields on a feature map that consistently stays the same spatial size as its input. The authors show that this does not only speed up the crack-segmentation process but also achieves comparable results when compared against larger architectures with much more parameters.

\textit{Loss Functions:}
Loss functions are a core component of DL methods.
Several works for crack segmentation have therefore focused on exploiting certain crack features, and designed specialised loss function \cite{kobayashi2018SpiralNetF1Based, mosinska2018PixelWiseLoss, li2021FastAccurate},

Common loss functions for supervised learning based crack segmentation include the cross entropy, focal loss \cite{lin2017FocalLoss}, Dice loss or combinations thereof.
The cross entropy loss for crack segmentation is often used as a binary cross entropy loss $L_{BCE}$ given that only the background and crack classes are available \cite{cheng2018PixelLevelCrack, konig2019SegmentationSurfacea, augustauskas2020AggregationPixelWise}. It measures the difference between the two probability distributions of the prediction $\hat{y}$,  $\hat{y} \in [0,1]$, and the ground truth $y$:
\begin{equation}
    L_{BCE}(y, \hat{y}) = y\log{\hat{y}} + (1-y)\log{(1-\hat{y})}
\end{equation}
One difficulty with the cross entropy loss is that the background and crack classes are weighted equally. One way to address this issue is to introduce a weighted term which weights the impact of the classes as introduced in the original U-Net \cite{ronneberger2015UNetConvolutional}. Another way is to use the focal loss, which has also found its application within the crack domain \cite{liu2019ComputerVisionBased, ren2020ImageBasedConcrete, qiao2021AutomaticPixelLevel}. Focal loss addresses the class imbalance issue by down-weighting easy examples, meaning pixels that a model can predict relatively well. Those easy examples can skew the performance of a model using basic losses such as cross entropy, as for cracks a large number of pixels belong to the background. To do this, the focal loss introduces a hyperparameter $\gamma$, which adjusts how easy examples are weighted. The standard recommended setting in \cite{lin2017FocalLoss} is $\gamma{=}2$. 
To derive the focal loss, the binary cross entropy loss can be rewritten as
\begin{equation}
    L_{BCE}(y, \hat{y}) = \begin{cases}
-\log{(\hat{y})}, & \text{for} \ y{=}1\\
-\log{(1-\hat{y})}, & \text{otherwise.}
\end{cases}
\end{equation}
This can then lead to further rewriting of the loss to $L_{BCE}(y, \hat{y}) = L_{BCE}(p_t) = -\log(p_t)$ if we define $p_t$ to be the class probability as follows:
\begin{equation}
    p_t = \begin{cases}
    \hat{y},  & \text{for} \ y{=}1\\
    1-\hat{y},  & \text{otherwise.}
    \end{cases}
\end{equation}
Focal loss then includes the $\gamma$ hyperparameter for tuning the weighting of easy examples and adds an optional class weight $\alpha$, $\alpha \in {[0,1]}$ to derive the equation as follows:
\begin{equation}
    L_{Focal} = -\alpha(1-p_t)^\gamma \log(p_t)
\end{equation}
The $\alpha$ hyperparameter is similar to the class weight in weighted cross-entropy losses and should be set higher for the less common class. 

The Dice overlap can be used as a performance metric (\autoref{sec:metrics}) but it is also used for model training as a loss function for cracks \cite{konig2021OptimizedDeep, konig2019ConvolutionalNeurala, konig2019SegmentationSurfacea, ghosh2021CrackWebmodified, augustauskas2020ImprovedPixelLevel}. 
Proposed in \cite{milletari2016VNetFully} it can be written as:
\begin{equation}
    L_{Dice}(y, \hat{y}) = 1-\frac{2\sum_i^N y_i\hat{y}_i + 1}{\sum_i^N y_i + \sum_i^N \hat{y}_i + 1}
\end{equation}
with $i$ being each individual pixel within a ground truth or prediction that have a total number of $N$ pixels. 

A specialized loss function based on the F1 metric is introduced in \cite{kobayashi2018SpiralNetF1Based}. This metric is often used to compare the performance of methods in computer vision tasks (also see \autoref{sec:metrics}). The proposed loss function, $L_{F1}$, is shown to outperform cross-entropy and focal loss functions when using multiple architectures:
\begin{equation}
    L_{F1}(y, \hat{y}) = -\log{\frac{2\sum^N_{i|y_i{=}1}\hat{y_i}}{N_{Crack} + \sum^N_{i}\hat{y_i}}}
\end{equation}
Here, $N_{Crack}$ stands for the total number of crack pixels and the term within the logarithm mirrors a differentiable version of the F1 metric. 

In \cite{li2021FastAccurate}, an exponentially weighted cross entropy loss function is proposed. It is shown to decrease training time as well as increase the results in comparison to using the standard cross entropy. Considering the weighted binary cross entropy, using a weighted term $w$:
\begin{equation}
 L_{wBCE}(y, \hat{y}) = wy\log{\hat{y}} + (1-y)\log{(1-\hat{y})}
\end{equation}
the weighted term is defined as $w=q(\alpha)=\alpha/(1-\alpha)$, where alpha refers to the ratio of crack to non crack pixels in a single batch of data. The authors from \cite{li2021FastAccurate} then found that setting this term to $q(\alpha) = \beta*10^{2\alpha-1}$ is optimal for the setting of crack segmentation. However, the choice of the hyperparameter $\beta, 0 < \beta < 1$ still remains.
Another loss function that leverages prior information of curvilinear structures such as cracks was introduced in \cite{mosinska2018PixelWiseLoss}. Here it is assumed that those structures follow a certain kind of topology, meaning that they are unlikely to have gaps or structures that are not of a curvilinear shape. This is implemented by computing the L2 distance of the feature maps created by interim layers of a pretrained VGG network $V$, after passing through the prediction and actual ground truth. This topological loss $L_{Top}$ can simplified as:
\begin{equation}
    L_{Top}(y, \hat{y}) = \sum^M_{m=1}\frac{1}{C} \sum^C_{c=1} || V^m_c(y) - V^m_c(\hat{y}) ||_2^2
\end{equation}
where $V^m_c$ denotes the activation output of the pretrained network at layer $m$ of all layers $M$ in the network and $c$ standing for the individual channel of layer $m$ with a total of $C$ channels.  

This work shows that the proposed loss function, in combination with an iterative refinement approach in which the image is passed through the same network multiple times, achieves increased performance with fewer topological errors. 

\textit{Other Approaches:} There are also many works for crack segmentation who do not focus on generating extra performance by designing complex architectures; but rather focus on  pooling layers \cite{konig2019SegmentationSurfacea, zhou2021MixedPooling, xu2021PushingEnvelope, han2021CrackWNetNovel}, leveraging prior information from the datasets \cite{xu2021PushingEnvelope} or using generative methods \cite{gao2019GenerativeAdversarial, zhang2020CrackGANPavement}.

Pooling layers are important parts of many fully convolutional segmentation architectures as they are responsible for spatial dimensional reduction through summarising regional features. As cracks can contain fine details, the work in \cite{konig2019SegmentationSurfacea} proposes a pooling layer for cracks that fuses the output of the basic maximum pooling operation, as well as the stacked pooling operation \cite{huang2018StackedPooling}, which aids in scale invariance. Another pooling function for cracks, introduced in \cite{zhou2021MixedPooling}, uses a fused output of performing average pooling using a horizontal and a vertical window with a width of 1 pixel to capture crack features better. Both of those methods were then shown to improve the crack segmentation performance by changing the pooling layer.
In \cite{xu2021PushingEnvelope} several methods are introduced that can increase the crack segmentation performance. They show that polynomial pooling \cite{wei2019BuildingDetailSensitive}, a function that balances max and average pooling, adds a slight performance gain in the crack segmentation task. Additionally, they also propose a novel post-processing step to remove small, erroneous shapes in crack segmentation outputs. For this, a variable autoencoder learns the shape prior of cracks and how to remove short or standalone segments that are not connected with the actual cracks. 

Another kind of approach which has shown to be successful within this task is to use ensemble models. Ensemble models combine the output of several models and often show better generalization capabilities, as each model might have learned to focus on distinctive features. One such method is shown within the work of Fan \textit{et al.} \cite{fan2020EnsembleDeep}. Based on their previous network architecture in \cite{fan2018AutomaticPavementa}, they propose to expand this by training the same model multiple times and merging their predicted output, which was shown to increase performance. Another approach to creating an ensemble is presented in \cite{wang2020NeuralNetwork}. Here, three U-Net-based architectures are first pretrained on crack data captures on dams, bridges, and walls. Each of those models is then fine-tuned on crack data from roads and tested on road data as well. In the first instance, the authors showed that transfer learning increased performance. After that, they ensemble all three of those models and were able to further increase the results on the road data. 

Generative adversarial networks \cite{goodfellow2014GenerativeAdversarial} have also been used for crack segmentation. In \cite{gao2019GenerativeAdversarial}, a modified U-net is used as a generative model and the discriminator of the GAN is used to gauge whether the generated segmentation map is ``real" in comparison with the actual segmentation label. 
Zhang \textit{et al.} \cite{zhang2020CrackGANPavement} propose a similar approach with a U-net-shaped generator and a discriminator who encourages the generator to produce coherent segmentation maps. However, in this approach, they also supplement the discriminator training by feeding manually created crack curves as ``real" images through it which limits overfitting. 
A novel approach incorporating reinforcement learning is proposed in \cite{park2021CrackDetection}. This two-stage approach uses a basic U-Net architecture before adding a refinement stage. The refinement stage is based on an Advantage Actor-Critic \cite{mnih2016AsynchronousMethods} approach and incorporates an agent that iteratively improves the initial segmentation map and closes gaps in the predictions.
Recent trends in the computer vision domain have discovered that transformers \cite{vaswani2017AttentionAlla}, a technique used previously for natural language processing, have immense potential in increasing the performance of vision algorithms \cite{dosovitskiy2021ImageWorth}. This also seems to be a promising approach for crack segmentation and was implemented by Guo \textit{et al.} \cite{guo2021TransformerBased}. Here, interim outputs of a standard segmentation CNN are split into patches, flattened to one dimension, and passed through a transformer based architecture to refine initial prediction results.

\subsubsection{Supervised Quantification Analysis}
Several works that perform supervised detection or segmentation of cracks also opt to predict measurements of the various dimensions of a crack.
In \cite{tong2017RecognitionAsphalt}, a crack is detected using a patch classification approach, followed by classifying each patch into a certain size range from 0-8 cm in 1cm steps. However, in this work, not the full crack is considered but only parts of it, due to the patch-based approach. \cite{yang2018AutomaticPixelLevel} propose an FCN approach to create a segmentation map and then use morphological operations to create a skeleton map of that crack. This skeleton is then used to calculate the pixel length of the crack, which could then be converted to the actual length when interpolating between the resolution and distance between the surface and camera. This is similar to the proposed approach in \cite{fan2020EnsembleDeep} where the crack skeleton is used for both, width and length calculation. In \cite{kim2019ImageBasedConcrete}, morphological operations are also used to quantify cracks but here they only propose to calculate the crack width. The calculation of crack width is achieved by computing the distance from the center (skeleton) axis of a crack to its outer edge. They also obtain the distance to the surface and use it with the camera sensors specifications as well as the pinhole camera model to compute the width of a crack in mm. 
Another approach in \cite{li2019AutomaticPixelLevel} computes the total surface area of a crack. This is achieved by using a laser range finder to obtain the distance to the surface, which is then used with a generated crack segmentation map using SegNet for the surface computation. In \cite{tran2020OneStage} a specialized vehicle with cameras is used to capture road surfaces with a resolution of 1mm per pixel. A RetinaNet \cite{lin2017FocalLoss} method is used to create bounding boxes for different instances of cracks and classify them as to whether they are fatigue cracks, longitudinal or transverse cracks. In this case, the cracks are labeled as longitudinal cracks if they are parallel or angling up to 45$^\circ$ to the road centerline, whereas longitudinal cracks have an angle greater than 45$^\circ$. The bounding box diagonal is then used to obtain the crack length.

\subsection{Semi and Weakly Supervised Learning}
In semi and weakly supervised learning, the available data is not as precise as for supervised learning. Weakly supervised annotations are usually not as accurate as they can contain noise or only consist of rough annotations. For the task of crack segmentation, this could mean that not all parts of a crack are correctly annotated and some crack regions might be missing annotations. In semi-supervised learning, only some of the data is sufficiently labeled but one also has access to a larger number of unlabeled data samples. 

To the best of the authors' knowledge, there is no research on semi and weakly supervised crack classification and detection. The works for this task mainly focuses on segmentation and quantification. Nonetheless, the approach shown in \autoref{fig:splitcrack} can be interpreted as a semi-supervised detection approach. Here the algorithm is only trained on classification labels and a previously unseen image can be split into patches, which are then classifying, and upon stitching those patches together a coarse detection map can be created. 

\subsubsection{Semi and Weakly Supervised Crack Segmentation}
\begin{figure} 
    \centering
    \captionsetup[subfigure]{labelformat=empty}
    \subfloat{\includegraphics[height=10em, width=0.55\linewidth]{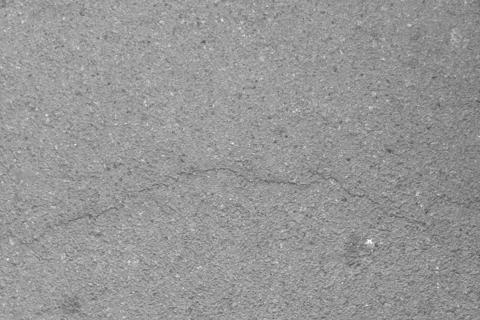}} \hspace{0.001\linewidth}
    \subfloat{\includegraphics[height=10em, width=0.43\linewidth]{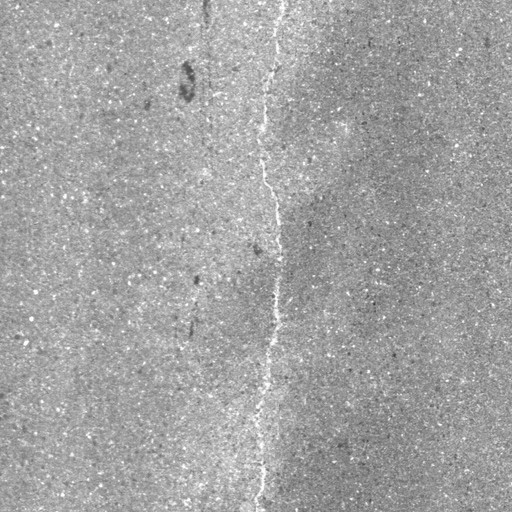}} \\ \vspace{-0.03\linewidth}
    \subfloat{\includegraphics[height=10em, width=0.55\linewidth]{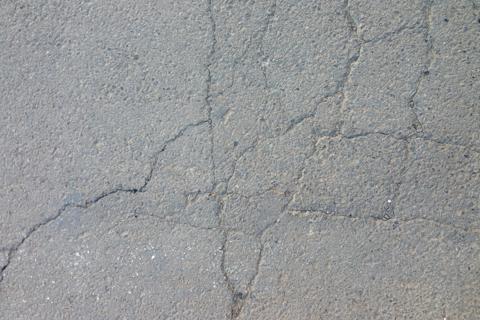}} \hspace{0.001\linewidth}
    \subfloat{\includegraphics[height=10em, width=0.43\linewidth]{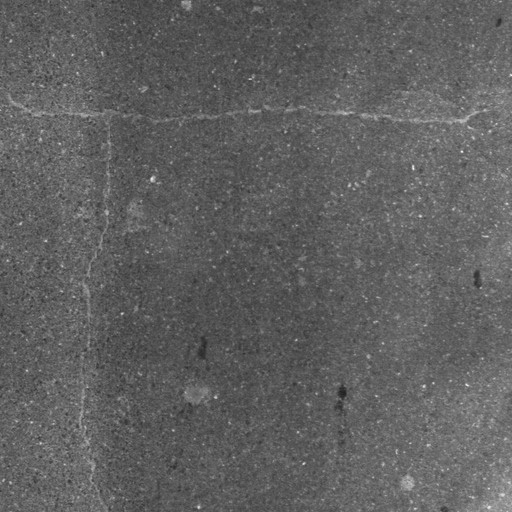}}\\
        \caption{Crack examples in which some of crack regions are not significantly darker than the background. Those ``edge cases" prove to be difficult examples for both, traditional and deep learning based methods.}
    \label{fig:difficult_cracks}
\end{figure}

In \cite{dong2020PatchBasedWeakly} a weakly supervised crack segmentation method is proposed. They split a segmentation labeled dataset into patches and manually label the image patches that have a crack. Using this, a classification network is then trained fully supervised to classify each patch as to whether it contains a crack or not. After training, each of those patches is passed through the classification network and a class activation map is generated which yields a rough crack-localization. This is followed by implementing a conditional random field to turn the class activation maps into synthetic labels. Using those synthetic, binary-segmentation labels a fully convolutional segmentation network is trained to predict cracks. 
The work in \cite{inoue2020CrackDetection} performs weakly supervised crack segmentation by reducing the quality of labels through a dilation process and add an additional component that can be added to any crack segmentation approach to increase performance. This second component is not DL based but rather exploits the crack characteristic that they are darker than the usual structure background. This component simply calculates the brightness of each pixel in relation to all other pixels in an image and darker pixels are assigned higher values. This brightness map is then multiplied with the predicted segmentation output and improves performance, even if the label quality during training was inaccurate. 
Fan \textit{et al.} \cite{fan2019RoadCrack} propose a semi-supervised approach that leverages the power of DL algorithms in combination with traditional methods for crack segmentation. They propose a two-stage approach in which a CNN for classification is trained fully supervised, which is then followed by applying the more classical techniques of image filtering and thresholding to patches that are of the crack class. This then yields a segmentation map as darker pixels are segmented as crack pixels. A similar approach is shown in \cite{konig2021WeaklySupervisedSurface}, where a classifier is used to create a detection map as shown in \autoref{fig:splitcrack}, which is then merged with a thresholding approach. Merging the detection map with the thresholding in this approach aims to suppress noise and wrongly segmented regions outside the actual crack area. The aforementioned approaches may work to a certain extent, however, in edge cases cracks may not always appear as darker regions within images which hinders segmentation performance. Some samples of this are shown in \autoref{fig:difficult_cracks}. Unfortunately, this still seems to be a problem for fully supervised segmentation methods as well \cite{zou2018DeepCrackLearning, konig2021OptimizedDeep}.

\subsubsection{Semi and Weakly Supervised Crack Quantification Analysis}
Inoue and Nagayoshi \cite{inoue2019DeploymentConsciousa} propose a semi-supervised way to learn the orientation of cracks. They rotate an input multiple times using different angles and feed them to a model during inference. This model has been trained using a custom loss function to only generate a high segmentation confidence score if the crack lies within a certain rotational angle. Using the given rotation angle of the input patch and the output of the model with the highest confidence that the center pixel of the input patch is a crack, they are able to determine the rotational crack angle. This is a semi-supervised approach as the input has not been annotated with crack rotation angles and due to its patch-based approach yields an output map in which each pixel of a crack input has been assigned a crack angle. 

\subsection{Unsupervised Learning}
In unsupervised learning, no labels are given to the algorithm. The aim is that the algorithm finds the patterns for itself. In the space of surface cracks with DL, limited work has been performed.
Chow \textit{et al.} \cite{chow2020AnomalyDetection} propose an unsupervised approach using an autoencoder. As a majority of images of surfaces do not contain anomalies such as cracks, they use the autoencoder to recreate the initial input image. If the recreation differs significantly from the input, it is assumed that a surface anomaly is present. They show that this approach fares well against other ML based approaches but do not provide a comparison against semi or fully supervised approaches.
Another unsupervised approach is proposed in \cite{yu2020UnsupervisedPixelLevel}. This uses an adversarial learning approach and transforms images into the frequency domain using a generator and then reverses this by reusing the generator weights to reverse the image from the frequency domain back to an image. By comparing this generated image with the actual image, this work is then able to detect the surface defects effectively. The whole approach is trained only on images that do not have cracks and therefore segments cracks and other anomalies using the reconstructed image.

\section{Practical Applications}
Many works do not only focus on the separate development of specific DL algorithms but also propose full systems or ML pipelines that either integrate software, multiple algorithms, and/or hardware.
This section highlights a selection of current approaches that also utilize DL based algorithms. 

A system to recognize cracks and leaks in tunnels is proposed in \cite{huang2018DeepLearning}. It consists of an image acquisition system that is maneuvered through tunnels and generates overlapping images. The captured images are then stitched together and a fully convolutional neural network takes on the task of segmenting the surface anomalies.
To enable more accurate quantification of cracks, the work in \cite{park2020ConcreteCrack} proposes to not only use a DL architecture to create detection bounding boxes around cracks but also include an auxiliary measurement system next to the image capturing equipment. This measurement system projects calibrated laser beams onto the surface which is captured using a camera. Using those laser points and the measured distance to the surface, which has also been captured, they can calculate a conversion between pixels and size in millimeters. This is then applied to the detected crack and the actual height and width can be determined. 
Jang \textit{et al.} \cite{jang2021AutomatedCrack}, propose a ring-shaped robot equipped with cameras to climb bridge piers and capture image data. The images are then stitched and preprocessed before passing them onto a modified SegNet algorithm to create segmentation maps. As the camera distance to the surface is also known, they propose to quantify the cracks using the euclidean distance transform and the pinpoint camera formula. 
In \cite{yuan2021NovelIntelligent} a remote-controlled robot equipped with stereo vision is used to capture surfaces of reinforced concrete. Using the data from stereo vision, a 3-D representation of the surfaces is created. The images are then fed through a Mask R-CNN \cite{he2017MaskRCNN} detection and instance segmentation algorithm to extract crack regions. This is followed by projecting the crack regions back into the 3-D space which then enables to view 3-D reconstruction of the surfaces with augmented and highlighted surface cracks.  

The applications discussed show that there is a wide range of potential applications that integrate DL algorithms for recognizing cracks in structural surfaces and the selection of algorithms needs to be made based on the task, data and labels available, and processing requirements. 

\section{Datasets and Performance Evaluation}
\label{sec:data_and_performance}
 The previous sections highlighted the DL algorithms and their applications within the field of SHM and surface cracks. In this section, the datasets and metrics that are used to gauge the performance of the algorithms are discussed and evaluated.
 
\subsection{Popular Open Datasets}
\label{sec:datasets}

 \begin{table*}[ht]
\renewcommand{\arraystretch}{1.2}
\centering
\caption{Overview over commonly used openly available surface crack datasets. Image-level annotations (\textit{I}) mean that a class label is attached to each image, detection annotation (\textit{D}) contain bounding boxes around objects such as cracks and pixel-level annotations (\textit{P}) have each pixel annotated to belonging to a certain class.}
\label{tab:datasets}
\begin{minipage}{\linewidth}
\begin{tabularx}{\linewidth}{lllXll}
\toprule
\multicolumn{1}{c}{\textbf{Reference}} &
\multicolumn{1}{c}{\textbf{Name}} &
\multicolumn{1}{c}{\textbf{Num. Images and Sizes}} &
\multicolumn{1}{c}{\textbf{Comment}} &
\multicolumn{1}{c}{\textbf{Annotations}} &
\multicolumn{1}{c}{\textbf{Train/Val/Test Split}} \\ \midrule
2011, \cite{zou2012CrackTreeAutomatic} & CrackTree & 200, ${800\times600}$px & Pavement images with occlusions such as shadows & \textit{P} & None\\
 \hline
2016, \cite{amhaz2016AutomaticCrack} & \textit{Multiple}\footnote{This publication consists of multiple datasets: AIGLERN, TEMPEST2, LRIS, LCMS and ESAR. Due to the small size of the datasets we have combined them.} & 66, various sizes & Road images from 5 different acquisition systems & \textit{P} & None \\ \hline
2016, \cite{shi2016AutomaticRoad} & CFD & 118, ${320\times480}$px & Road surfaces in Beijing & \textit{P} & None \\ \hline
2017, \cite{eisenbach2017HowGet}& GAPs v1 & 1,969, ${1920\times1080}$px & Data captured by specialized vehicle with multiple damage types such as cracks and potholes & \textit{D} & 1,418 / 51 / 500\\ \hline
2018, \cite{ozgenel2018PerformanceComparison}& CCIC & 40k, ${227\times227}$px & Images of concrete with optional cracks & \textit{I} & None\\ \hline
2018, \cite{dorafshan2018SDNET2018Annotated}& SDNet2018 & 56k, ${256\times256}$px & Images of pavements, walls and bridge decks with optional cracks & \textit{I} & None\\ \hline
\multirow{ 4}{*}{2018, \cite{zou2018DeepCrackLearning}} 
    & CrackTree260 & 260, various sizes &  Expansion of the CrackTree \cite{zou2012CrackTreeAutomatic} dataset  & \textit{P} & 260 / - / - \\
    & Stone331 & 331, ${512\times512}$px &  Stone surfaces & \textit{P} & - / - / 331 \\
    & CRKWH100 & 100, ${512\times512}$px &  Road pavements, captured under laser illumination & \textit{P} & - / - / 100 \\
    & CrackLS315 & 315, ${512\times512}$px &  Road pavements, captured under laser illumination & \textit{P} & - / - / 315 \\\hline
2019, \cite{liu2019DeepCrackDeepa} & DeepCrack & 537, $544\times384$px &  Diverse surface textures and scenes & \textit{P} & 300 / - / 237 \\\hline
\multirow{ 2}{*}{2019, \cite{yang2019FeaturePyramid}} 
    & Crack500 & 500, ${2000\times1500}$ & Pavement cracks, captured with smartphone & \textit{P} & 250 / 50 / 200\\
    & GAPs384 & 384, ${1920\times1080}$ & Subset of 384 crack images from GAPS v1 \cite{eisenbach2017HowGet} & \textit{P} & / - / - / 384\\ \hline
\multirow{ 2}{*}{2019, \cite{stricker2019ImprovingVisual}} &
\multirow{ 2}{*}{GAPs v2} & 2,468, various sizes & Expansion of GAPs v1\cite{eisenbach2017HowGet} & \textit{I},\textit{D},\textit{P} & 1,417 / 551 / 500\\
 & & 50k, various sizes & Subset of GAPS v2 for classification & \textit{I} & 50k / 10k / 10k\\
\bottomrule
\end{tabularx}
\end{minipage}
\end{table*}
 
A common problem, similar to many other fields utilizing DL, is the availability of relevant datasets. The thorough labeling of crack data is a difficult and time-consuming task and there are only a limited amount of datasets available. 
There is also a wide variation in the quality of datasets, some authors download freely available images online and label them themselves \cite{tang2019MultiTaskEnhanced, liu2019DeepCrackDeepa}, some use specialized tools for data collection and use experts to annotate them \cite{chambon2011AutomaticRoad, eisenbach2017HowGet}.
\autoref{tab:datasets} highlights some of the most common datasets used within this space. It also contains the number of images within this dataset, the train-test-val split of the data (if available), and their annotation quality. 
\subsubsection{Issues}
One of the issues with the datasets is the general availability of the data. Many authors use datasets but do not publish them for open access which can make evaluations and comparison of different algorithms difficult. . Another issue with the available datasets is the lack of a separate validation set;
during the training process, after each epoch, one usually measures the generalization capability of a method by studying their performance on the separate validation set. An often-used method in DL is to select the weights of a model where the best validation performance is achieved. However, without having a specific validation set, authors can run into several problems. In one instance, they might need to split the already limited training data into train and validation sets, often without knowing if and how other works have split their data. This also can lead to less available training data, which can negatively impact the algorithm performance. In contrast to that, without actually using a validation set, algorithms weights may be selected that have not trained to a sufficient fit, meaning they may be severely under or overfitted to the training data. 
Another issue, with some of the early datasets which are still used, is that the authors have not defined a data split \cite{chambon2011AutomaticRoad, zou2012CrackTreeAutomatic, shi2016AutomaticRoad}. This can lead to some further works using a random or set split- with other works implementing their own split. The result of that is that a direct comparison between algorithms is often not possible, even though the same dataset was used. 

\begin{figure}
	\centering
	\captionsetup[subfigure]{labelformat=empty}
	\subfloat[]{\includegraphics[width=0.8\linewidth, ]{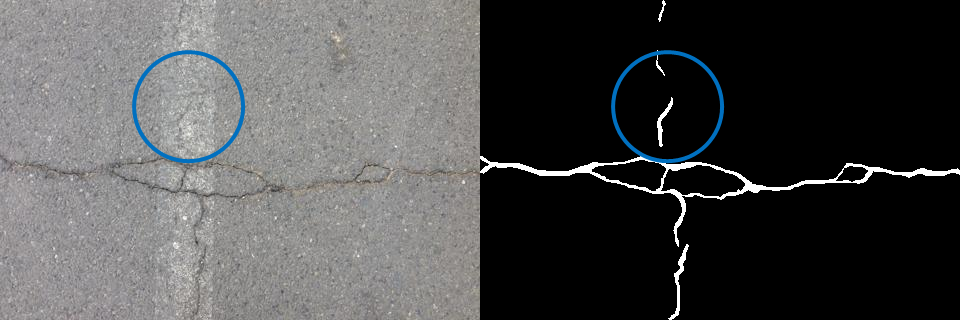}} \vspace{-2em}
	\subfloat[Missing/Inaccurate Annotations]{\includegraphics[width=0.8\linewidth, ]{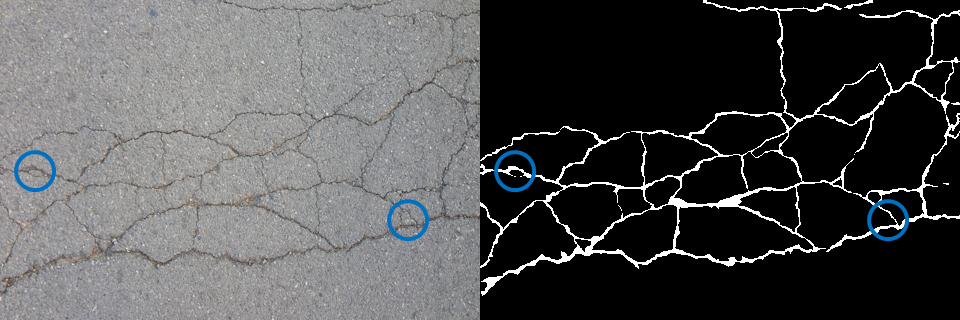}} \\
	\subfloat[Annotations too thin]{\includegraphics[width=0.44\linewidth, ]{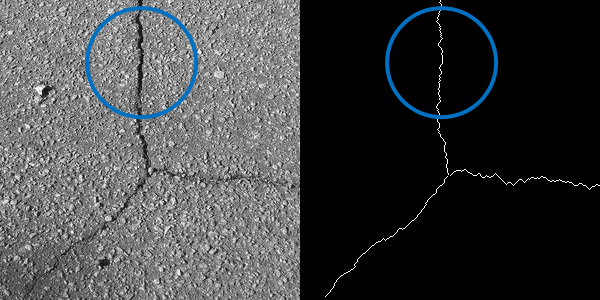}} \hspace{0.05\linewidth}
	\subfloat[]{\includegraphics[width=0.44\linewidth, ]{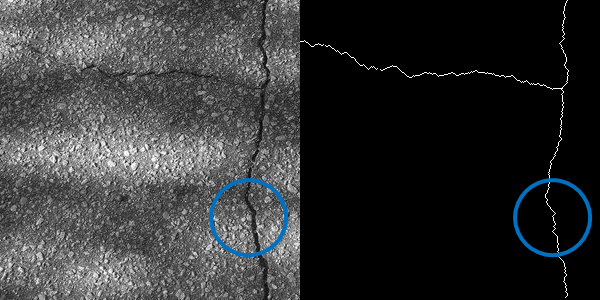}} \\
	\subfloat[Annotations too thick]{\includegraphics[width=0.44\linewidth, ]{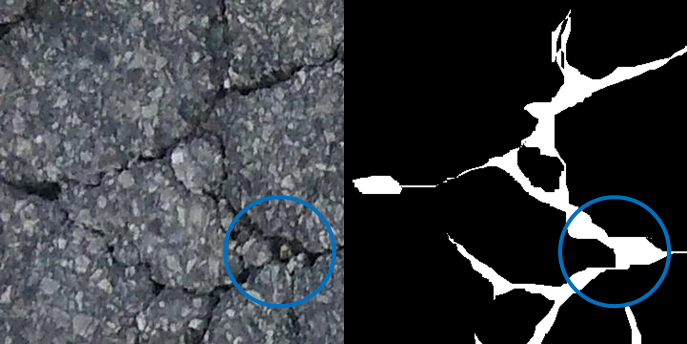}} \hspace{0.05\linewidth}
	\subfloat[]{\includegraphics[width=0.44\linewidth, ]{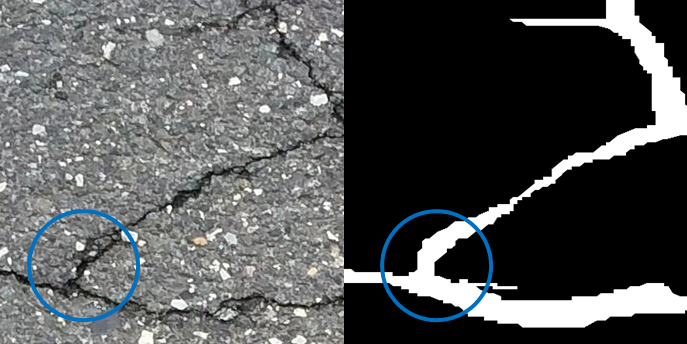}}\\
	\caption[Examples of crack images and their segmentation labels which are inaccurate.]{Examples of crack images and their segmentation labels which are inaccurate. Two samples are shown for each of the following datasets in that order: CFD \cite{shi2016AutomaticRoad}, CT260 \cite{zou2018DeepCrackLearning}, DeepCrack \cite{liu2019DeepCrackDeepa}}
	\label{fig:wrong_labels}
\end{figure}
\subsubsection{Label Quality}
\label{sec:label_qual}
Labeling crack data can be a difficult, time-consuming and tedious process that is subjective to human error \cite{oliveira2009AutomaticRoad}. Depending on the type of label, thorough labeling of a single image can take up to several minutes \cite{inoue2020CrackDetection}. This can be an expensive process as a specialist might have a high hourly charge and using crowd-sourcing to label images such as in ImageNet may exceed the initial budget. To work around this, some authors have labeled images themselves, without having stated that they have sufficient expertise within this field \cite{shi2016AutomaticRoad, zou2018DeepCrackLearning, liu2019DeepCrackDeepa}. On the other hand, some works have worked in conjunction with experts or companies that are specialized within this domain to generate the datasets \cite{chambon2011AutomaticRoad, eisenbach2017HowGet}.
In some data, it may be obvious where cracks are but some other data may have hairline cracks or structural features (which are not cracks) but can be labeled as cracks. \autoref{fig:wrong_labels} shows some examples where label annotations appear to be incorrect or lacking detail. To overcome the issue of inaccurate labeling within the crack segmentation task, some works have used a relaxed metric that takes label inaccuracies into account when scoring the performance. This is further outlined in \autoref{sec:metrics}.

\begin{figure}
    \centering
    \captionsetup[subfigure]{labelformat=empty}
    \subfloat[Original]{\includegraphics[width=0.3\linewidth, ]{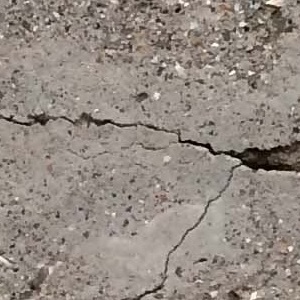}} \hspace{0.01\linewidth}
    \subfloat[Brightness Change]{\includegraphics[width=0.3\linewidth]{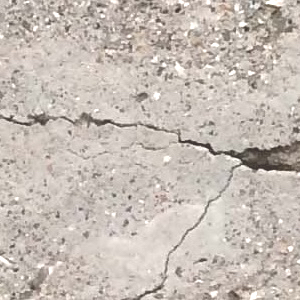}} \hspace{0.01\linewidth}
    \subfloat[Contrast Change]{\includegraphics[width=0.3\linewidth]{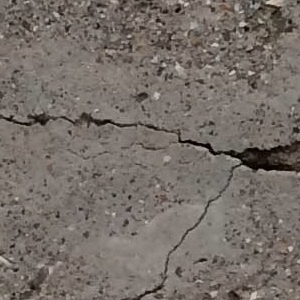}} \\ \vspace{-0.03\linewidth}
    \subfloat[Gamma Change]{\includegraphics[width=0.3\linewidth]{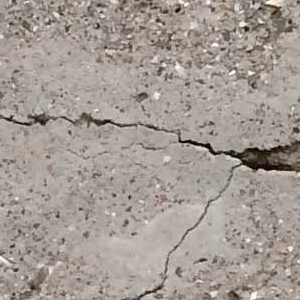}} \hspace{0.01\linewidth}
    \subfloat[Saturation Change]{\includegraphics[width=0.3\linewidth]{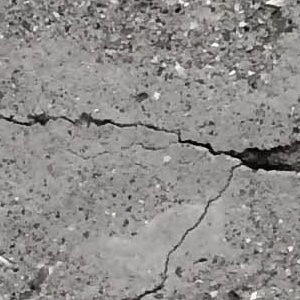}} \hspace{0.01\linewidth}
    \subfloat[Scale Change]{\includegraphics[width=0.3\linewidth]{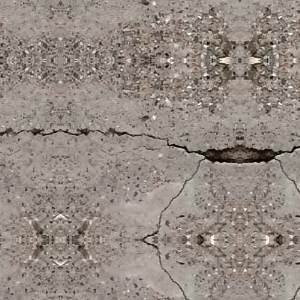}}
    \caption{Example image augmentations of a crack image showing the ability to create a larger number of samples by applying augmentations.
    }
    \label{fig:augmentation}
\end{figure}
\subsubsection{Augmented and Artificial Data}
Due to the relatively small number of training images available, it is common to attempt to increase the available training data through augmentation. This is a basic approach, which is very popular in the DL domain are image augmentations. Those augmentation operations are randomly applied to images during training and can contain operations that change the brightness, contrast, or color of an image or include flips and rotations. Those flips along the horizontal and vertical axis, as well as rotations, are popular for cracks \cite{zou2018DeepCrackLearning, liu2019DeepCrackDeepa, konig2021OptimizedDeep, xu2021automatic}, as they are usually rotation invariant. This does not apply to several other domains; for example, flipping images of numbers upside down may result in incorrect representations.
Augmentation can not only be applied during training. The work in \cite{konig2021OptimizedDeep} uses test-time-augmentation to average the predictions of their method after feeding the same image through the network multiple times after it has been resized. This is shown to consistently increase performance.
Another recent approach that gained popularity within the image classification domain on ImageNet is to automatically learn the augmentation strategies to best fit the training dataset \cite{cubuk2018AutoaugmentLearning, hataya2019FasterAutoaugment}. However, to the best of the authors' knowledge, at the time of writing this review, this has not been attempted within the DL domain for cracks.
Patch-based approaches can also be used to increase the available training data. For example, in \cite{jenkins2018DeepConvolutionala, konig2019ConvolutionalNeurala, konig2019SegmentationSurfacea} a sliding window is used to extract patches from the CFD dataset, which increases the available training samples exponentially.
Synthesizing images and adding them to the body of available training data can also be a viable approach in increasing data. Whilst this has been shown to work very well within other domains \cite{wang2018HighResolutionImage, zhou2020HiNetHybridFusion}, little research has been performed for cracks. In \cite{mazzini2020NovelApproach} a data generative approach is proposed, which first creates labels followed by the creation of images. Initially, a semantic segmentation layout (segmentation label) with classes such as road marking and cracks is artificially generated. This is followed by loading data from the main dataset which follows a similar layout of the generated one. At last, both the similar data and the layout are passed through a synthesizer which then creates synthetic images that are used to supplement the available training data. This has shown to increase performance, especially in classes that were underrepresented in the original dataset.

\subsection{Performance Evaluation Metrics}
\label{sec:metrics}
Depending on the task such as detection or classification, different metrics are used to evaluate the performance of DL based algorithms for cracks. 
Let $TP$, $FP$ be a true-positive and false-positive prediction and $TN$, $FN$ a true negative and false negative prediction respectively. One can then, depending on the task, use a $TP$ prediction as a correctly predicted image/patch (classification) or a correctly predicted pixel (segmentation). 
\subsubsection{Common Metrics}
Using those predictions, some of the most common metrics for cracks are precision $PR$, recall $RE$, and the $F1$ score. The $F1$ score is based on the harmonic mean of recall and precision. 
\begin{equation}
    PR =  \frac{TP}{TP + FP}
\end{equation}
\begin{equation}
    RE = \frac{TP}{TP + FN}
\end{equation}
\begin{equation}
    F1=2\cdot\frac{PR\cdot RE}{PR+RE}
\end{equation}
To illustrate those metrics, crack semantic segmentation recall states how much of the actual crack pixels have been correctly predicted by the method and precision as to how many of the predicted pixels are actually on a crack. 
A large number of works use those metrics to report their results for classification \cite{zhang2016RoadCracka, pauly2017DeeperNetworksa, wang2017GridBasedPavementa, riid2019PavementDistress}, detection \cite{tang2019MultiTaskEnhanced, carr2018RoadCracka} or segmentation \cite{cheng2018PixelLevelCrack, jenkins2018DeepConvolutionala, konig2019ConvolutionalNeurala, inoue2019DeploymentConsciousa}. 
It is common to use $PR$, $RE$, and $F1$ on boolean data, meaning the results are either true (1) or false (0). This is achieved by thresholding results at a specific prediction confidence $t$, with $0<=t<=1$, with all results below that threshold being false and all above, true such as in  \cite{fan2018AutomaticPavementa, konig2019ConvolutionalNeurala, augustauskas2020ImprovedPixelLevel, lau2020AutomatedPavement, xu2021PushingEnvelope}.

Accuracy $ACC$ is another metric that is often used to show the performance. Results are often reported with classification accuracy or per-pixel segmentation accuracy. 
\begin{equation}
    ACC=\frac{TP+TN}{TP+TN+FP+FN}
\end{equation}
However, unlike previous metrics, $ACC$ is not well suited for reporting results with a high class imbalance. Whilst some image classification datasets are evenly distributed with crack or non-crack images, for segmentation or detection in most datasets only a small fraction of an image belongs to a crack. If an image only has 1\% of crack pixels and a segmentation algorithm predicts all pixels to be of the background class it would still achieve a 99\% accuracy. Despite that, several works using crack datasets make use of it to compare their results \cite{pauly2017DeeperNetworksa, tong2017RecognitionAsphalt, liu2019DeepCrackDeepa, riid2019PavementDistress, augustauskas2020AggregationPixelWise}.

The Intersection over Union metric ($IoU$) is another metric that scores the overlap between a prediction and a ground truth. It is often used for detection and segmentation tasks \cite{tang2019MultiTaskEnhanced, dong2020PatchBasedWeakly, louk2020PavementDefect, liu2019DeepCrackDeepa}. Let $X$ be a predicted output and $Y$ the ground truth, then $IoU$ is calculated as follows:
\begin{equation}
    IoU=\frac{|X \cap Y|}{|X \cup Y|}
\end{equation}
The IoU is also referred to as mean-IoU in the literature, however as a majority of the datasets using cracks have binary labels (crack or no-crack), this part is omitted in the equation.

Less common as a performance metric, but still used for reporting crack segmentation performance is the Dice metric \cite{wu2020CrackDetecting, augustauskas2020ImprovedPixelLevel, sun2020RoadwayCracka}. This metric is calculated very similar to the $F1$ score, with the difference that here the confidence scores in the range of $[0,1]$ are being used. Therefore it can be written as:
\begin{equation}
    Dice= 2\cdot\frac{|X \cap Y|}{|X| + |Y|}
\end{equation}

For segmentation and detection tasks it is possible to apply the aforementioned metrics either on the whole testing dataset at once or aggregate them after calculating the results on a per image level. In this case, the $TP$, $FP$, $FN$, and $FP$ are either accumulated across the whole datasets, and the metric is calculated or it is calculated on each image following by averaging the results. However, whilst many authors report what metric they used, this point is often not addressed. 

\subsubsection{Other Metrics}
Recently, several other metrics have also been used, or have specially been created for the various crack tasks;
Optimal image scale ($OIS$) and optimal dataset scale ($ODS$) are two metrics reused from works performing edge-detection \cite{xie2015HolisticallyNestedEdge} and have found their adaptation for measuring crack segmentation performance \cite{zou2018DeepCrackLearning, yang2019FeaturePyramid, guo2021BARNetBoundary, konig2021OptimizedDeep, park2021CrackDetection}. 
\begin{equation}
    OIS =  \frac{1}{N_{img}}  \sum_{i}^{N_{img}}  max \Big\{ F1_t^i: \forall t \in \{0.01, ..., 0.99\} \Big\}
\end{equation}
\begin{equation}
    ODS = max \Big\{ \big\{ \frac{1}{N_{img}}  \sum_{i}^{N_{img}}   F1_t^i \big\} : \forall t \in \{0.01, ..., 0.99\} \Big\}
\end{equation}
In those equations, $N_{img}$ stands for the number of images available in the testing dataset, $i$ a single image and $t$ delimits the confidence threshold for a pixel belonging to the class crack. Essentially, this uses an oracle, which has access to the ground truth labels, to determine the confidence threshold that generated the best result of a $F1$ score on the image $OIS$, or dataset $ODS$ scale.
Another metric for this task, which we have coined cumulative $ODS$ ($cODS$) was also used within works for crack segmentation \cite{liu2019DeepCrackDeepa, cheng2018PixelLevelCrack, konig2021OptimizedDeep}. Here, the $F1$ score is calculated after summing all $TP$, $FP$, $FN$ pixels of a dataset $d$, and then the best possible confidence threshold is selected that yields the highest $F1$ score.
\begin{equation}
    cODS = max \Big\{ F1_t(d) : \forall t \in \{0.01, ..., 0.99\} \Big\}
\end{equation}
Another segmentation metric, proposed in \cite{yang2019FeaturePyramid} is $AIU$. $AIU$ computes the average of all IoUs for each confidence threshold in range $[0,1]$:
\begin{equation}
    AIU = \frac{1}{T} \sum_t^T \frac{|X_t \cap Y|}{|X_t \cup Y|}
\end{equation}
with T being the total number of thresholds used.

Other metrics that have also been used is the area under the precision-recall curve reported as the average precision \cite{zou2018DeepCrackLearning, tang2019MultiTaskEnhanced, kobayashi2018SpiralNetF1Based}, Macro-F1 \cite{inoue2020CrackDetection}, as well as the Matthews correlation coefficient \cite{riid2019PavementDistress}.

\subsubsection{Issues}
A common metric issue within the field of computer vision tasks on surface cracks is that works that use the same datasets opt to use different metrics to report results. This makes comparisons of the performance of algorithms difficult. 
Nearly all of the classification metrics are straightforward and do not appear to have issues, however, for segmentation in this space, there are several issues. 
Some works in segmentation resort to skeletonization of the cracks before calculation of the metric \cite{yang2019FeaturePyramid, guo2021BARNetBoundary}. Whilst in itself this is sufficient for tasks such as edge detection, it may not accurately represent the actual performance of algorithms; The resulting metrics of a skeletonized prediction only of the centerline of a crack compared against the sekletonized prediction of all crack pixels are going to be highly similar, even though the latter would yield a better coverage of the crack.
It is also not often reported how the predicted segmentation results have been thresholded to create a binary segmentation map. Some metrics alleviate that ($OIS$, $ODS$, $cODS$, $AIU$) but the standard F1 or IoU metric do not have such a mechanism built-in. Therefore it can be unclear at what threshold those results have been generated such as in \cite{inoue2019DeploymentConsciousa, dong2020PatchBasedWeakly, qu2020CrackDetection}.
Another issue for segmentation is the label quality which leads to ``loosening" of the metric in several works \cite{zou2018DeepCrackLearning, fan2018AutomaticPavementa, inoue2019DeploymentConsciousa, jenkins2018DeepConvolutionala, konig2021OptimizedDeep}. In these works, $TP$ pixels have an ``allowance" of being within a two-pixel or five-pixel distance to the ground-truth map to be still counted as a $TP$ instead of being classed as a $FP$. Through the way metrics such as $F1$ are calculated, this can lead to oversaturating the number of $TP$ pixels by simply creating wider predictions, which leads to better results in the metrics but objectively worse results when manually evaluating the segmentation maps. 

\section{Research Gaps and Directions}
In this section, we outline several of the research gaps that we have identified within this review. Those research gaps, when addressed, can show potential for future work tackling the SHM tasks for cracks.
\subsection{Datasets}
Whilst several datasets for the various tasks showing cracks are available, there is a lack of large scale openly-available datasets. In the general computer vision domain with DL, the ImageNet dataset, with its substantial amount of samples, has accelerated DL research. Therefore it would be very beneficial for this domain to also have a single, large-scale dataset that is used to train and evaluate algorithms. The GAPs v2 subset in \cite{stricker2019ImprovingVisual} and SDNet2018 \cite{dorafshan2018SDNET2018Annotated} with over 50k images for classification are steps in the right direction but it would also be beneficial to have annotated data for the segmentation and quantification tasks. 
A simple solution for this would be to collate several of the datasets such as proposed in \cite{drouyer2020AllTerrain}. However, there still is the issue with incorrect labels, which spans across several datasets in this space as outlined in \autoref{sec:label_qual}. It might be useful for this data to be relabeled by a suitably experienced group of experts so that the label quality is consistent and correct. Unfortunately, this would incur a large number of costs and time. 

\subsection{Metrics}
The large amount of work carried out in this field has led to a variety of different evaluation procedures, datasets, and metrics are being used. Work that carries out similar tasks in this field would majorly benefit from a unified evaluation procedure and well-defined metrics which would severely increase the comparability of algorithms. This can be handled by authors providing a detailed, step-by-step description of measures taken to handle predictions, equations of the metrics used, and evaluation procedures in their work which would increase reproducible results. Alternatively, authors could provide their source code which evaluates the predictions.

In terms of what metrics being used, we recommend that the loosening of segmentation metrics as described in \autoref{sec:metrics} should be avoided. This does not seem to be a sufficient alternative to mitigate the quality of badly annotated data and can promote algorithms that have a worse real-world performance by predicting segmentation maps too ``thick". Additionally, it would be beneficial to always outline whether a fixed confidence threshold cutoff has been used and also provide measures in alternative metrics that do not incorporate only binary predictions. 

Due to the growing amount of research interest for crack quantification, it may also be valuable to define a common, standardized metric for measuring the correctness of prediction degrees of rotation within cracks or their various spatial dimensions such as length and thickness. 

\subsection{Semi, Weakly and Unsupervised Learning}
In comparison to the number of works within the supervised learning space for cracks, the semi, weakly and unsupervised learning approaches are underrepresented. 
Here, similar to the previously mentioned dataset issues, it may be beneficial to create a standardized dataset for those types of learning. Current approaches have authors changing the label quality \cite{inoue2020CrackDetection, dong2020PatchBasedWeakly} of given datasets manually, which may not be the most optimal way going forward. A unified dataset for those types of learning can accelerate research and provide better comparability of given algorithms. 
The number of unsupervised learning approaches that incorporate DL in this space is still very small. Especially due to the difficulties in obtaining and correctly labeling data, research within this area could prove to be very valuable for automatic SHM processes.

\subsection{Temporal Data}
Another issue, which extends well beyond the DL algorithms, is the availability of temporal data showing cracks. The datasets currently available only show cracks at a single point in time. Whilst specialized are able to tell from their expertise the severity and future development of cracks, there is no dataset available yet that contains data that shows the development of cracks over a certain time period. 
The prediction of sequential data has made significant progress due to DL in other domains such as video sequence prediction \cite{guen2020DisentanglingPhysical} or time series forecasting \cite{lim2021TimeSeriesForecasting}. SHM could significantly benefit from such datasets as it would support predictive maintenance measures. DL algorithms could be applied to predict the propagation of cracks at a later time-point and determine if and when precautionary maintenance measures need to be applied. However, the limitations for this are similar to the generation of normal datasets. In addition, cracks can take time to grow and the creation of such dataset with, train, validation, and test data may take several years. 

\section{Conclusion}
Currently, DL algorithms achieve state-of-the-art results in a multitude of fields and have also found their applications in SHM. 
This review presented an overview of DL algorithms that are used within the space of crack classification, detection, and quantification within the domain of SHM using computer vision. Specifically, we reviewed approaches using different types of learning: supervised, semi-supervised, weakly-supervised and unsupervised, together with an outline of the common metrics and datasets that are used. We outline the issues faced by research within this field, deliver an overview of potential gaps in the research, and deliver a view on directions for future research.
As a main result, we have identified the abundance of research within the supervised learning domain but note the difficulty in comparing architectures and performance due to non-standardized common metrics, datasets, and issues within dataset annotations. In the future, we hope that those issues will be addressed and believe that more work within the semi, weakly and unsupervised domain within this field will push the research forward and alleviate the issues that come with small datasets and difficult annotations.

\IEEEpeerreviewmaketitle

\appendices

\ifCLASSOPTIONcaptionsoff
  \newpage
\fi

\bibliographystyle{IEEEtran}
\bibliography{ref}


\end{document}